\documentclass[final,3p,times,twocolumn]{elsarticle}

\usepackage{amssymb}
\usepackage{amsmath}

\usepackage{xcolor}
\usepackage{microtype}
\usepackage{graphicx}
\usepackage{subfigure}
\usepackage{booktabs}
\usepackage{hyperref}
\usepackage{algorithm}
\usepackage[noend]{algorithmic}

\usepackage{amsmath}
\usepackage{amssymb}
\usepackage{mathtools}
\usepackage{amsthm}
\usepackage{multirow}
\usepackage{threeparttable}
\usepackage{textcomp}
\usepackage{stfloats}
\usepackage{color}
\usepackage[utf8]{inputenc}
\usepackage{tabularx}
\usepackage{graphicx}
\usepackage{arydshln}
\usepackage{hyperref, varioref}
\usepackage[capitalize,noabbrev]{cleveref}
\theoremstyle{plain}

\theoremstyle{definition}

\theoremstyle{remark}

\journal{  }

\begin{document}

\begin{frontmatter}



\title{Adaptive Substructure-Aware Expert Model for Molecular Property Prediction} 




\author[1,2]{Tianyi~Jiang\fnref{equal}}
\ead{josieyi0319@163.com}

\author[1,2]{Zeyu~Wang\fnref{equal}}
\author[1,2]{Shanqing~Yu}
\author[1,2]{Qi~Xuan\corref{cor1}}
\ead{xuanqi@zjut.edu.cn}

\affiliation[1]{organization={Institute of Cyberspace Security, College of Information Engineering},
                addressline={Zhejiang University of Technology}, 
                city={Hangzhou},
                postcode={310023}, 
                state={Zhejiang Province},
                country={China}}

\affiliation[2]{organization={Binjiang Institute of Artificial Intelligence},
                addressline={Zhejiang University of Technology},
                city={Hangzhou},
                postcode={310056}, 
                state={Zhejiang Province},
                country={China}}

\fntext[equal]{Tianyi Jiang and Zeyu Wang contribute equally to this manuscript.}
\cortext[cor1]{Corresponding author.}


\begin{abstract}
Molecular property prediction is essential for applications such as drug discovery and toxicity assessment. While Graph Neural Networks (GNNs) have shown promising results by modeling molecules as molecular graphs, their reliance on data-driven learning limits their ability to generalize, particularly in the presence of data imbalance and diverse molecular substructures. Existing methods often overlook the varying contributions of different substructures to molecular properties, treating them uniformly. To address these challenges, we propose ASE-Mol, a novel GNN-based framework that leverages a Mixture-of-Experts (MoE) approach for molecular property prediction. ASE-Mol incorporates BRICS decomposition and significant substructure awareness to dynamically identify positive and negative substructures. By integrating a MoE architecture, it reduces the adverse impact of negative motifs while improving adaptability to positive motifs. Experimental results on eight benchmark datasets demonstrate that ASE-Mol achieves state-of-the-art performance, with significant improvements in both accuracy and interpretability.
\end{abstract}




\end{frontmatter}


\section{Introduction}\label{sec:introduction}

In recent years, with the rapid development of artificial intelligence, molecular property prediction has gradually become a hot research topic in fields such as drug discovery~\cite{walters2020applications}. This technology can significantly reduce the cost and risk associated with traditional wet-lab drug development~\cite{wang2024multi}. Among various molecular property prediction techniques, deep learning methods based on Graph Neural Networks~\cite{wang2024sub} have garnered significant attention due to their ability to perceive molecular topological structures. Specifically, these methods primarily model molecules as molecular graphs and transform molecular property prediction tasks into graph-level classification or regression problems. However, due to the data-driven nature of these methods~\cite{jiang2024mix, wang2024know, wang2023null}, their reliance on high-quality labeled molecular data and the issue of data imbalance significantly limit the generalization ability of these models, thereby lacking practical application value.

\begin{figure}[!t]
    \centering
    \includegraphics[width=\linewidth]{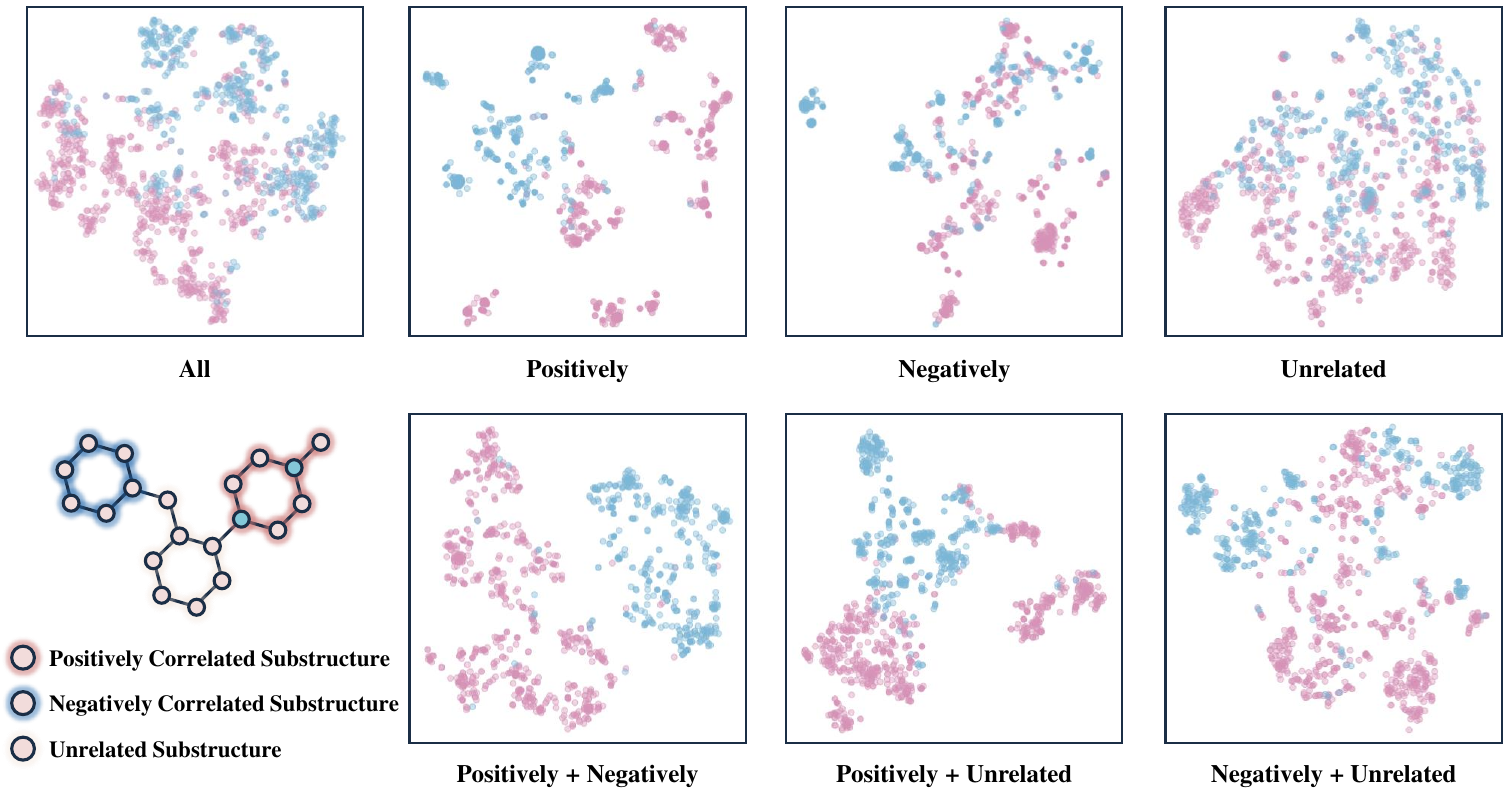}
    \caption{The T-SNE visualization on different substructures.}
    \label{fig: introduce}
\end{figure}

To address these issues, researchers have introduced the concept of multi-expert models, which aim to improve model performance by integrating predictions from multiple expert models~\cite{6215056}. Specifically, this approach decomposes complex problems into multiple sub-problems, allowing each expert model to focus on different subsets and thereby enhancing the generalization ability of the overall model. In molecular property prediction tasks, researchers emphasize the deep exploration of molecular topological information and the incorporation of multi-expert techniques to tackle the challenges of generalization to complex molecular topologies. For example, GraphDIVE~\cite{ijcai2022p289} proposed a global and local multi-level expert learning framework to address the impact of imbalanced data on model performance; GMoE~\cite{wang2024graph} focused on subgraph and motif-level features and designed a dynamic information aggregation strategy for multi-expert systems. Although these methods have achieved good results in practice, they often overlook domain-specific features of molecular topologies. To address this gap, TopExpert~\cite{kim2023learning} proposed a domain-specific expert model based on molecular scaffolds. By leveraging the high correlation between molecular scaffolds and molecular properties, this model further improved performance. However, in molecular structures, factors closely related to molecular properties are not limited to molecular scaffolds but also include a variety of functional group structures. Moreover, different substructures may have varying degrees of influence and contribution to molecular properties. To further explore the relationship between molecular substructures and molecular properties, this paper conducts clustering visualization experiments using structural features at different levels. Specifically, this paper adopts Breaking of Retrosynthetically Interesting Chemical Substructures (BRICS)~\cite{degen2008art} decomposition to split molecules and analyzes the impact of each substructure on prediction performance. On this basis, molecular substructures are categorized into three types: those positively correlated with molecular properties, those negatively correlated, and those unrelated, corresponding to the most beneficial, most detrimental, and remaining substructures, respectively. From the experimental results shown in~\cref{fig: introduce}, it can be observed that in the single-angle substructure feature view, the T-SNE projection of positively and negatively correlated substructures exhibits relatively clear clustering structures, indicating that these substructures are strongly correlated with molecular properties. In contrast, the projection of uncorrelated substructures exhibits more dispersed clustering patterns with blurred classification boundaries. While in the combined multi-angle feature view, the joint view of positively and negatively correlated substructures enables a clearer separation of different molecular properties, further enhancing the discriminative ability of the model. Based on these experimental observations, we propose that substructures with clearer classification boundaries are more strongly correlated with molecular properties, while others are weakly correlated. Furthermore, the results from the joint view further confirm that molecular properties are influenced by multiple substructures with varying degrees of correlation.

In conclusion, to effectively address the generalization issues caused by imbalanced and insufficient data in existing models, this paper proposes ASE-Mol, a novel MoE framework that enhances molecular representation learning by dynamically identifying and leveraging positive and negative substructures. ASE-Mol uses BRICS decomposition to fragment molecules into key substructures and classifies them into positive and negative motifs based on their impact on molecular property prediction through attribution analysis. The model assigns separate expert networks to process positive and negative motifs, allowing each expert to specialize in specific structural patterns. Additionally, the framework employs two routers: the positive motif router determines routing scores based on motif embeddings and a learnable weight matrix, and the negative motif router combines embeddings from both positive and negative motifs to attenuate the adverse effects of negative motifs. The final prediction is obtained by combining the outputs from these expert models. Extensive experiments on benchmark datasets validate the effectiveness of ASE-Mol. Our contributions are summarized as follows: 
\begin{itemize}
    \item We propose ASE-Mol, a novel MoE framework for molecular property prediction. By dynamically adapting to different molecular substructures, the framework avoids the confusion that arises when a single model processes diverse and complex molecular features.
    \item ASE-Mol leverages BRICS decomposition and essential substructure awareness to identify and distinguish between positive and negative substructures. By combining these insights, the model achieves a more comprehensive understanding of molecular properties.
    \item Leveraging a MoE architecture, ASE-Mol employs a dynamic routing mechanism to assign distinct molecular substructures to specialized experts, enabling effective processing and integration of both positive and negative substructures.
    \item Extensive experiments show the effectiveness of ASE-Mol. In addition, interpretability analysis confirms that the method pinpoints key substructures that influence molecular properties, providing valuable insights into the molecular features that drive specific properties.
\end{itemize}

\section{Related Work}\label{sec:related work}

\subsection{Molecular Property Prediction}

Molecular property prediction is a fundamental task in cheminformatics and bioinformatics, essential for applications such as drug discovery and toxicity assessment~\cite{walters2020applications, li2022deep}. Traditional approaches often rely on hand-crafted molecular descriptors combined with machine learning models like random forests or support vector machines~\cite{xia2023understanding}. These approaches, while effective to some extent, are limited by their reliance on predefined feature sets. Recent advances have shifted focus toward graph-based representations, where molecules are modeled as graphs with atoms as nodes and bonds as edges. This paradigm has driven the development of GNNs, which achieve state-of-the-art performance by iteratively aggregating information from neighboring nodes and edges to learn rich molecular representations~\cite{wang2023null, xuan2019subgraph, zhou2020m, yang2024state}. However, despite their success, GNN-based methods struggle with generalization because they often treat molecular graphs as undifferentiated wholes, overlooking the distinct roles and contributions of different substructures. This limitation reduces the ability of the model to capture structure-property relationships effectively. To address this limitation, recent studies have explored methods such as substructure masking and functional group-based embeddings~\cite{zhang2021motif, xie2023self, liu2024rethinking}, which aim to enhance interpretability by aligning molecular representations with chemical intuition. These approaches emphasize the pivotal role of specific molecular substructures in determining overall molecular properties. By incorporating these substructures, models can provide a more complete understanding of the molecular properties.

\subsection{Expert Models}

The MoE framework has emerged as a promising approach for dynamically allocating computational resources based on input characteristics~\cite{shazeer2017}. This architecture consists of multiple specialized sub-models, or ``experts'', each designed to capture specific aspects of the input data, with a gating mechanism determining the contribution of each expert~\cite{jacobs1991adaptive, ma2018modeling}. MoE frameworks have been successfully applied to graph-based tasks, enabling adaptive expert selection and improving performance by routing graphs to specific experts based on structural patterns~\cite{wang2024graph, zhang2024graph, yao2024moe}. Within molecular property prediction, expert-based models have demonstrated the ability to capture finer-grained structure information~\cite{gaspar2022glolloc, wu2022alipsol, zhang2024mol}. For instance, GraphDIVE~\cite{ijcai2022p289} utilizes weighted combinations of expert outputs to improve graph classification tasks, while TopExpert~\cite{kim2023learning} employs scaffold alignment strategies to enhance both interpretability and accuracy. However, existing methods often overlook the need to explicitly handle both positive and negative substructures, which play distinct roles in determining molecular properties. Building upon these works, our approach integrates a substructure-aware expert module aiming to better integrate the contributions of these substructures.

\begin{figure*}[t]
    \centering
    \includegraphics[width=\linewidth]{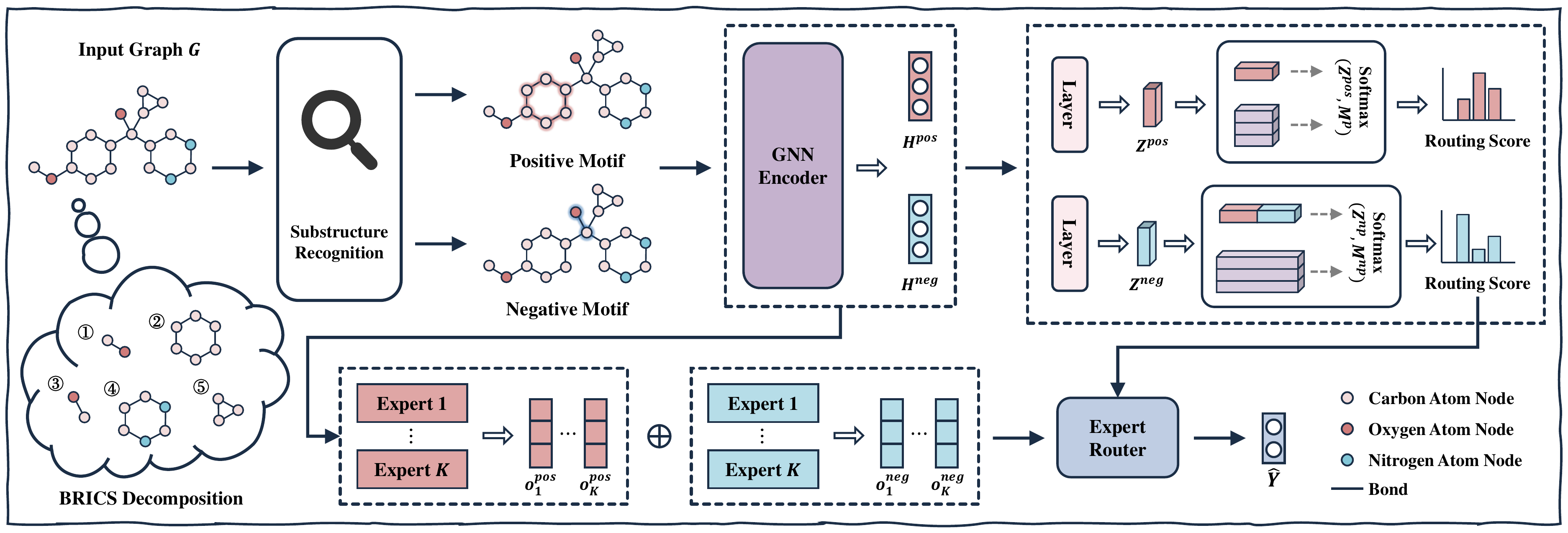}
    \caption{Overview of the proposed ASE-Mol.}
    \label{fig: framework}
\end{figure*}

\section{Preliminaries}\label{sec:preliminaries}

\subsection{Problem Definition}

Let a graph $G$ represent a molecule, where nodes correspond to atoms and edges correspond to chemical bonds between atoms. The graph is characterized by a node features $X\in \mathbb{R}^{N\times d}$ and an adjacency matrix $A\in \mathbb{R}^{N\times N}$, where $N$ is the number of atoms and $d$ is the dimensionality of the node features. If an edge exists between nodes $u$ and $v$, $A(u,v)=1$, and $A(u,v)=0$ otherwise. Given a training set $\mathcal{D}=\{(G_i, Y_i)\}^B_{i=1}$, where $G_i$ denotes a molecular graph and $Y_i\in \{0, 1\}^T$ represents the true labels for $T$ molecular properties, our task is to learn a mapping function $f: G\to Y$ that predicts the molecular properties. Specifically, the objective is to learn molecular embeddings for $B$ molecules in the dataset, enabling accurate property prediction. 

\subsection{Graph Neural Networks}

To encode useful information from graph-structured data, GNNs leverage a message-passing mechanism that iteratively computes node representations by aggregating information from neighboring nodes and edges in a layer-wise manner. The representation of a node $u$ at the $(l+1)$-th GNN layer is given by:
\begin{equation}
    \label{GNN}
    h_u^{(l+1)} = \text{GNN}\big(h_u^{(l)}, \{h_v^{(l)}, e_{uv}|v\in \mathcal{N}(u)\}\big)
\end{equation}
where $\mathcal{N}(u)$ represents the set of neighboring nodes of $u$ and $e_{uv}$ represents the edge feature between nodes $u$ and $v$. Then, the graph-level representation $H$ is computed by applying a readout function, which can be described as:
\begin{equation}
    \label{readout}
    H = \text{READOUT}(\{h^l_u|u\in G\})
\end{equation}
where the readout function $\text{READOUT}(\cdot)$ can be implemented using various pooling functions, such as summing pooling, averaging pooling, max pooling, or attention pooling. This graph-level representation $H$ is typically used for the molecular property prediction task.

\subsection{Mixture of Experts}

The MoE framework comprises a set of expert networks $E=\{E_k\}^K_{k=1}$, each with its own trainable parameters, and a gating network $Q(x)$ that dynamically selects and combines expert outputs based on the input $x$. The gating network assigns a set of scores $Q(x)=\{q_k(x)\}^K_{k=1}$, which determine the contribution of each expert to the final output. 

For a given input feature $x$, the output of the MoE module is computed as a weighted sum of expert outputs:

\begin{equation}
    \label{expert_output}
    \hat{Y} = \sum^K_{k=1} q_k(x) \cdot E_k(x)
\end{equation}
where $E_k(x)$ represents the output of the $k$-th expert. By dynamically routing inputs to specialized experts, the MoE framework enhances model efficiency while maintaining adaptability to diverse input distributions.

\section{Methods}\label{sec:methods}

In this section, we introduce ASE-Mol, a GNN-based framework for molecular property prediction that employs a MoE approach. We begin by introducing the identification and learning of molecular substructures, followed by proposing a substructure-based MoE module that dynamically selects and combines expert outputs. An overview of ASE-Mol is depicted in~\cref{fig: framework}. 

\subsection{Molecular Substructure Recognition and Learning}

\subsubsection{BRICS Recognition}

To analyze the contribution of distinct molecular substructures, we first adopt the BRICS approach to fragment the molecule $G$ into a collection of essential substructures $G^{sub} = \{g^{sub}_j\}^S_{j=1}$. This method effectively isolates specific substructures from various molecules.

The key idea of substructure recognition is to generate masks that selectively hide certain fragments from a GNN model. By evaluating the impact of these unmasked substructures on the model, we can determine which fragments have a significant impact on the prediction results. If these substructures can improve the prediction accuracy, they are classified as ``positive'', otherwise they are labeled as ``negative''.

Specifically, inspired by SME~\cite{wu2023chemistry}, the GNN model predicts molecular properties through a Multi-Layer Perceptron (MLP) layer. To achieve this, we first compute the corresponding substructure embedding, defined as follows:
\begin{equation}
    \label{mask}
    mask_{g^{sub}_j} = 
    \left\{
    \begin{aligned}
        &1, \text{if node $u$ is in the substructure} \\
        &0, \text{if node $u$ is not in the substructure} \\
    \end{aligned}
    \right.
\end{equation}

\begin{equation}
    \label{mask_embedding}
    H^{sub}_j = mask_{g^{sub}_j} \cdot H^V
\end{equation}
where $H^V=\{h_u^{(l+1)}\}^N_{u=1}$ is the set of node representations of the molecule, $mask_{g^{sub}_j}$ is the masking vector of the substructure $g^{sub}_j$, and $H^{sub}_j$ is the graph embedding of it.

We refer to the influence of using only the substructures on the overall prediction performance as the attribution. To obtain the attribution of each substructure, we perform two predictions on the molecular graph: one before masking and one after masking. The difference between these predictions quantifies the attribution, which is formulated as:
\begin{equation}
    \label{attribution}
    \text{Attribution}_{g^{sub}_j} =
    \begin{cases}
    \text{MLP}(H^{sub}_j) - \text{MLP}(H)       & ,Y=1 \\
    \text{MLP}(H) - \text{MLP}(H^{sub}_j)       & ,Y=0 \\
    \end{cases}
\end{equation}

Then, we define the top-$\psi$ highest attribution substructures as the ``positive motif'' and the top-$\psi$ lowest as the ``negative motif'', subsequently obtaining their respective molecular representations $H^{pos}$ and $H^{neg}$. 

\subsubsection{Learning Optimization}

To optimize the recognition and learning of these substructures, we incorporate a margin triplet loss~\cite{schroff2015facenet} and a downstream task loss. 

We treat the original molecular graph, positive motif, and negative motif as three distinct views. The graph-level representations learned from these views should ensure high-quality, and the positive and the negative motif should be well-discriminated to some extent. Specifically, the margin triplet loss is denoted as:

\begin{equation}
    \label{margin_loss}
    \begin{aligned}
        \mathcal{L}_{margin} = \frac{1}{B} \sum_{i=1}^B \mathbb{E}(-\max(&\sigma(H_i, H_i^{pos}) - \sigma(H_i, H_i^{neg})+\epsilon, 0))
    \end{aligned}
\end{equation}
where $\sigma(\cdot, \cdot)$ is the sigmoid function and $\epsilon$ is a margin value. Then, we combine the task-relevant loss with $\mathcal{L}_{margin}$:
\begin{equation}
    \label{sub_loss}
    \mathcal{L}_{rec} = \mathcal{L}_{task} + \alpha \mathcal{L}_{margin}
\end{equation}
where $\alpha$ is a parameter used to balance the trade-off between $\mathcal{L}_{margin}$ and $\mathcal{L}_{task}$.

\subsection{Substructure-based MoE Module}

\subsubsection{Multiple Experts Representation}

Following substructure identification, the identified positive and negative substructures are grouped into multiple expert models. The objective of this part is to route the positive and negative substructures based on their representation and assign them to specialized expert models.

For $T$ tasks, we employ $TK$ experts for each type of substructures, where $K$ represents the number of experts. Each expert is responsible for capturing the distinct influence of specific structural patterns on molecular property prediction. Given the $k$-th expert $E_k$, the graph embedding computed by the expert is formulated as:
\begin{equation}
    \label{expert}
    E_k(H^{sub}) = H^{sub} W_k
\end{equation}
where $W_k \in \mathbb{R}^{TK\times d}$ is a weight matrix. For each type of expert, $H^{sub}$ is divided into $H^{pos}$ or $H^{neg}$.

\subsubsection{Substructure-based Expert Routing}

To improve the adaptiveness of these experts, we propose a mechanism to learn a dynamical routing score using a router. To be more specific, we route each graph based on its graph embedding $E_k(H^{sub})$. Here, we design two different routers for two motifs. 

For the ``positive motif'', the router is parameterized as a weight matrix $M^p\in \mathbb{R}^{K\times d}$. The ``positive motif'' router embedding $Z^{pos}$ is projected from a mapping layer $\text{Linear}(\cdot)$. Then, we compute the dot product with the router embedding and the weight matrix. The routing score $r_k^p$ can be calculated by:
\begin{equation}
    \label{routing p}
    r^p_k = \frac{\exp{({Z^{pos}}^\top M^p_k / \tau+\varepsilon)}}{\sum_{k'=1}^K \exp{({Z^{pos}}^\top M^p_{k'} / \tau+\varepsilon)}}
\end{equation}
where $\tau = 0.1$ is the temperature hyper-parameter, and $\varepsilon$ is sampled noise~\cite{riquelme2021scaling}.

For the ``negative motif'', representing a substructure negatively correlated with the target properties, we aim to mitigate its adverse impact on the optimization of the experts. To achieve this, we combine the embeddings from both positive and negative motifs, enabling the model to learn more predictive features from the negative motif through the routing mechanism. The router for this case is parameterized as $M^{np}\in \mathbb{R}^{K\times 2d}$. We concatenate both the ``positive motif'' router embedding $Z^{pos}$ and the ``negative motif'' router embedding $Z^{neg}$ to get a joint router embedding $Z^{np}$. The routing score $r_k^{np}$ is then calculated as:
\begin{equation}
    \label{joint router embedding}
    Z^{np} = [Z^{pos}, Z^{neg}]
\end{equation}

\begin{equation}
    \label{routing np}
    r^{np}_k = \frac{\exp{({Z^{np}}^\top M^{np}_k / \tau+\varepsilon)}}{\sum_{k'=1}^K \exp{({Z^{np}}^\top M^{np}_{k'} / \tau+\varepsilon)}}
\end{equation}
where $Z^{neg}$ is also projected from a mapping layer $\text{Linear}(\cdot)$. Then, the positive logit $o^{pos}$ and the negative logit $o^{neg}$ can be obtained by a combination of all experts according to the routing scores:

\begin{equation}
    \label{logit p}
    o^{pos} = \sum_{k=1}^K r^{p}_k E^{pos}_k
\end{equation}

\begin{equation}
    \label{logit np}
    o^{neg} = \sum_{k=1}^K r^{np}_k E^{neg}_k
\end{equation}

Finally, we compute the predicted value for each molecule as follows:

\begin{equation}
    \label{y}
    \hat{Y}_i = o^{pos}_i \oplus o^{neg}_i
\end{equation}
where $\oplus$ is the concatenation operation.

\subsubsection{Learning Optimization}

To optimize the expert learning process, we introduce the downstream task loss. Furthermore, we empirically observed that a few experts tend to dominate across all instances during training, a phenomenon previously reported in other MoE methods~\cite{shazeer2017, jiang2024mope}. To encourage better specialization among experts, we introduce an additional importance loss function to penalize dominant experts. For a given batch of graphs $\mathcal{B}=\{G_1,...,G_b\}$, the importance of each expert is defined as the summed routing score in the batch, $\text{Imp}(E_k)=\sum_{G_j \in \mathcal{B}} r_k$. The importance loss is calculated as the mean coefficient of variation of all expert importance values:

\begin{equation}
    \label{loss_imp}
    \mathcal{L}_{imp} = \sigma \big((\frac{\text{std}(\{\text{Imp}(E_k)\}^K_k)}{\text{mean}(\{\text{Imp}(E_k)\}^K_k)})^2; \gamma\big)
\end{equation}
where $\sigma(\cdot)$ is a stop-gradient function that prevents the error propagation of this loss term when the coefficient of variation is below a predefined threshold $\gamma=0.1$. So the parameters in ASE-Mol are optimized by minimizing the following loss:

\begin{equation}
    \label{loss_total}
    \mathcal{L}_{total} = \mathcal{L}_{task} + \beta \mathcal{L}_{imp} 
\end{equation}
where $\beta$ is the balance parameter that control the quality of two losses.

\begin{algorithm}[!t]
    \caption{Training procedure of ASE-Mol}
    \label{algorithm}
    \begin{algorithmic}[1]    
        \REQUIRE A training dataset $\mathcal{D}$, A GNN model $f_\theta$, An expert model $f_\eta$, Number of recognition steps $T_{rec}$, Number of total epochs $T_{total}$
        \ENSURE the GNN model parameters $\theta$ and the expert model parameters $\eta$
        \FOR{each graph $G\in \mathcal{D}$}
            \STATE Perform BRICS decomposition, split $G$ into $G^{sub} = \{g^{sub}_i\}^S_{i=1}$;
        \ENDFOR
        \STATE Model parameters initialization;
        \FOR{$e\in [0,T_{rec}]$}
            \FOR{each batch $\mathcal{B}\in \mathcal{D}$}
                \STATE Recognize Substructure via~\cref{attribution};
                \STATE Obtain $H^{pos}$ and $H^{neg}$ via $f_\theta$;
                \STATE Compute $\mathcal{L}_{rec}$ via~\cref{sub_loss};
                \STATE $f_\theta \gets f_\theta - \bigtriangledown_{f_\theta}(\mathcal{L}_{rec})$;
            \ENDFOR
        \ENDFOR
        \FOR{$e\in [0,T_{total}]$}
            \FOR{each batch $\mathcal{B}\in \mathcal{D}$}
                \STATE Obtain the graph embedding by the expert via~\cref{expert};
                \STATE Compute the routing score via~\cref{routing p} and~\cref{routing np};
                \STATE Obtain the predicted values via~\cref{y};
                \STATE Compute $\mathcal{L}_{total}$ via~\cref{loss_total};
                \STATE $f_\theta \gets f_\theta - \bigtriangledown_{f_\theta}(\mathcal{L}_{total})$, $f_\eta \gets f_\eta - \bigtriangledown_{f_\eta}(\mathcal{L}_{total})$;
            \ENDFOR
        \ENDFOR
    \end{algorithmic}
\end{algorithm}

\subsection{Training Procedure}

The training procedure of ASE-Mol is summarized in~\cref{algorithm}. During training, we first apply BRICS decomposition to split each graph into substructures. Next, we recognize the top-$\psi$ highest and lowest attributions to extract the ``positive motif'' and the ``negative motif''. This process is optimized using the loss $\mathcal{L}_{rec}$ and the number of iterations is adaptively adjusted by evaluating matches before and after each update. After motif recognition, we employ the substructure-based MoE module to generate predictive values and optimize the model by combining the task loss with the importance loss.

\section{Experiment}\label{sec:experiment}

In this section, we conduct comprehensive experiments to evaluate the performance of ASE-Mol in molecular property prediction. We first describe the experimental settings, followed by comparisons with MoE-based baselines. Next, we perform ablation studies to assess the contributions of each module. We further investigate the model’s interpretability by visualizing key substructures and router assignments. Finally, we conducted a hyper-parameter sensitivity and a time complexity analysis.


\subsection{Experiment Setup}

\subsubsection{Datasets}

We evaluated the proposed ASE-Mol on eight benchmark datasets from MoleculeNet~\cite{wu2018moleculenet}. Consistent with previous works, these datasets were split into training, validation, and test sets following a 80\%: 10\%: 10\% ratio, using the scaffold splitter from Chemprop~\cite{heid2023chemprop}. The statistics of the datasets are summarized in~\cref{tab: datasets}.

\begin{table}[!t]
    \caption{Statistics of datasets.}
    \centering
    \renewcommand{\arraystretch}{1.05}
    \resizebox{0.45\textwidth}{!}{
    \begin{tabular}{cccc}
    \toprule
    \textbf{Dataset} & \textbf{\# Compounds} & \textbf{\# Tasks} & \textbf{\# avg. BRICS} \\
    \midrule
    BBBP & 2039 & 1 & 4.07 \\
    ClinTox & 1478 & 2 & 4.93 \\
    HIV & 41127 & 1 & 4.14 \\
    SIDER & 1427 & 27 & 6.60 \\
    Tox21 & 7831 & 12 & 3.53 \\
    ToxCast & 8575 & 617 & 3.82 \\
    MUV & 93087 & 17 & 5.32 \\
    BACE & 1513 & 1 & 7.22 \\
    \bottomrule
    \label{tab: datasets}
    \end{tabular}
    }
\end{table}

\subsubsection{Baselines}

To demonstrate the effectiveness of our method, we compared ASE-Mol with four GNN backbone models, including GCN~\cite{kipf2017semisupervised}, GIN~\cite{xu2018how}, GAT~\cite{velivckovic2018graph}, and GraphSAGE~\cite{hamilton2017inductive}. And for a comprehensive comparison, we compare our approach with various baselines: MoE, E-Ensemble~\cite{dietterich2000ensemble}, GraphDIVE~\cite{ijcai2022p289}, and TopExpert~\cite{kim2023learning}. For all baselines, we leveraged the results reported in the previous study~\cite{kim2023learning} and conducted our experiments with the same hyper-parameters as those used in their models. The detailed descriptions of all baselines are provided below.

\begin{itemize}
    \item \textbf{Mixture of Experts (MoE)} employs a multi-layer perceptron with Gumbel-Softmax to select the most relevant experts for each molecule.
    \item \textbf{E-Ensemble}~\cite{dietterich2000ensemble} aggregates the outputs from the experts by calculating their arithmetic mean.
    \item \textbf{GraphDIVE}~\cite{ijcai2022p289} combines the expert outputs using a weighted sum, with the weights learned through a linear layer followed by a Softmax function.
    \item \textbf{TopExpert}~\cite{kim2023learning} utilizes a gating module along with a scaffold alignment strategy to compute a weighted sum of the outputs from experts.
\end{itemize}

\begin{table*}[!t]
    \caption{Performance across GNN backbones and eight benchmark datasets.}
    \centering
    \renewcommand{\arraystretch}{1.15}
    \resizebox{\textwidth}{!}{
    \begin{tabular}{c|c|cccccccc}
    \bottomrule
    \textbf{Encoder} & \textbf{Methods} & \textbf{BBBP} & \textbf{Tox21} & \textbf{ToxCast} & \textbf{SIDER} & \textbf{ClinTox} & \textbf{MUV} & \textbf{HIV} & \textbf{BACE} \\
    \hline 
    \multirow{6}{*}{GCN} & original & 65.9 $\pm$ 0.9 & 74.4 $\pm$ 0.6 & \underline{63.6 $\pm$ 1.1} & 60.6 $\pm$ 0.8 & 55.4 $\pm$ 3.6 & 74.0 $\pm$ 1.3 & 75.2 $\pm$ 1.4 & 71.0 $\pm$ 4.6 \\
    \cdashline{2-10}[1pt/1pt]
          & MoE & 65.4 $\pm$ 1.6 & 74.3 $\pm$ 0.4 & 61.5 $\pm$ 0.9 & 60.8 $\pm$ 1.0 & 68.1 $\pm$ 2.4 & 73.9 $\pm$ 1.2 & 75.5 $\pm$ 1.0 & 75.8 $\pm$ 2.8 \\
          & E-Ensemble & 65.8 $\pm$ 2.8 & 74.4 $\pm$ 0.6 & 61.6 $\pm$ 0.7 & 60.3 $\pm$ 0.9 & \underline{70.5 $\pm$ 5.9} & 74.1 $\pm$ 1.0 & 74.9 $\pm$ 1.0 & 76.0 $\pm$ 0.2 \\
          & GraphDIVE & 62.5 $\pm$ 1.9 & 73.2 $\pm$ 0.9 & 62.0 $\pm$ 0.7 & 51.9 $\pm$ 2.0 & 60.4 $\pm$ 6.2 & 64.6 $\pm$ 4.7 & 65.2 $\pm$ 4.1 & 59.4 $\pm$ 4.6 \\
          & TopExpert & \underline{67.0 $\pm$ 2.4} & \underline{75.3 $\pm$ 0.4} & 63.1 $\pm$ 0.6 & \underline{60.8 $\pm$ 0.9} & 70.1 $\pm$ 6.0 & \underline{74.6 $\pm$ 0.6} & \underline{76.6 $\pm$ 0.7} & \underline{76.4 $\pm$ 1.9} \\
    \cdashline{2-10}[1pt/1pt]
          & \textbf{ASE-Mol} & \textbf{77.5 $\pm$ 1.8} & \textbf{77.6 $\pm$ 0.8} & \textbf{73.1 $\pm$ 0.7} & \textbf{64.0 $\pm$ 1.2} & \textbf{83.2 $\pm$ 4.7} & \textbf{78.7 $\pm$ 1.4} & \textbf{85.5 $\pm$ 1.1} & \textbf{96.3 $\pm$ 1.2} \\
    \hline 
    \multirow{6}{*}{GIN} & original & 68.9 $\pm$ 1.2 & 74.3 $\pm$ 0.6 & \underline{64.1 $\pm$ 1.6} & 58.1 $\pm$ 1.5 & 58.8 $\pm$ 5.7 & 76.1 $\pm$ 1.3 & 75.6 $\pm$ 1.6 & 69.0 $\pm$ 4.7 \\
    \cdashline{2-10}[1pt/1pt]
          & MoE & 66.3 $\pm$ 2.0 & 74.5 $\pm$ 0.5 & 60.1 $\pm$ 1.0 & 58.6 $\pm$ 0.9 & 55.5 $\pm$ 3.0 & \underline{76.1 $\pm$ 0.8} & 71.4 $\pm$ 2.7 & 68.8 $\pm$ 3.9 \\
          & E-Ensemble & 66.5 $\pm$ 1.5 & 74.4 $\pm$ 1.1 & 60.7 $\pm$ 1.1 & 56.1 $\pm$ 1.6 & 59.8 $\pm$ 7.2 & 72.8 $\pm$ 2.5 & 76.2 $\pm$ 1.1 & 68.3 $\pm$ 5.2 \\
          & GraphDIVE & 65.0 $\pm$ 2.6 & 72.1 $\pm$ 3.0 & 54.7 $\pm$ 1.2 & 52.9 $\pm$ 2.3 & 52.9 $\pm$ 6.9 & 65.5 $\pm$ 7.0 & 68.9 $\pm$ 2.1 & 62.5 $\pm$ 4.7 \\
          & TopExpert & \underline{70.0 $\pm$ 0.7} & \underline{75.3 $\pm$ 0.7} & 62.6 $\pm$ 0.4 & \underline{58.9 $\pm$ 1.2} & \underline{60.3 $\pm$ 4.5} & 75.7 $\pm$ 1.6 & \underline{76.3 $\pm$ 1.4} & \underline{71.7 $\pm$ 4.0} \\
    \cdashline{2-10}[1pt/1pt]
          & \textbf{ASE-Mol} & \textbf{76.7 $\pm$ 0.8} & \textbf{80.3 $\pm$ 0.8} & \textbf{73.3 $\pm$ 1.1} & \textbf{66.5 $\pm$ 1.1} & \textbf{80.6 $\pm$ 4.3} & \textbf{85.6 $\pm$ 1.7} & \textbf{81.9 $\pm$ 1.7} & \textbf{91.0 $\pm$ 2.9} \\
    \hline 
    \multirow{6}{*}{GAT} & original & 64.9 $\pm$ 1.2 & 75.0 $\pm$ 0.8 & \underline{63.5 $\pm$ 1.6} & 61.0 $\pm$ 1.1 & 58.9 $\pm$ 1.4 & \underline{74.5 $\pm$ 0.9} & 75.5 $\pm$ 1.7 & 75.3 $\pm$ 2.4 \\
    \cdashline{2-10}[1pt/1pt]
          & MoE & 64.0 $\pm$ 1.4 & 70.8 $\pm$ 0.7 & 62.3 $\pm$ 0.8 & 60.0 $\pm$ 1.6 & 54.1 $\pm$ 4.7 & 73.1 $\pm$ 1.8 & 73.0 $\pm$ 1.9 & 76.5 $\pm$ 2.8 \\
          & E-Ensemble &\underline{ 66.8 $\pm$ 1.5} & 72.2 $\pm$ 1.2 & 62.5 $\pm$ 0.6 & 59.4 $\pm$ 4.1 & 58.7 $\pm$ 4.4 & 73.5 $\pm$ 1.5 & 75.2 $\pm$ 1.1 & \underline{77.0 $\pm$ 3.2} \\
          & GraphDIVE & 64.1 $\pm$ 1.4 & 70.1 $\pm$ 1.3 & 60.4 $\pm$ 1.3 & 53.7 $\pm$ 1.7 & \underline{60.2 $\pm$ 7.2} & 73.1 $\pm$ 1.6 & 75.5 $\pm$ 1.4 & 68.4 $\pm$ 7.5 \\
          & TopExpert & 65.4 $\pm$ 2.1 & \underline{74.9 $\pm$ 0.8} & 62.9 $\pm$ 0.9 & \underline{62.0 $\pm$ 1.3} & 59.1 $\pm$ 2.5 & 74.1 $\pm$ 1.1 & \underline{77.3 $\pm$ 1.3} & 76.3 $\pm$ 2.0 \\
    \cdashline{2-10}[1pt/1pt]
          & \textbf{ASE-Mol} & \textbf{82.4 $\pm$ 1.6} & \textbf{76.8 $\pm$ 0.7} & \textbf{74.6 $\pm$ 0.8} & \textbf{63.6 $\pm$ 1.1} & \textbf{87.8 $\pm$ 4.1} & \textbf{80.2 $\pm$ 1.7} & \textbf{86.3 $\pm$ 2.3} & \textbf{94.9 $\pm$ 2.0} \\
    \hline 
    \multirow{6}{*}{GraphSAGE} & original & \underline{68.1 $\pm$ 1.5} & 74.2 $\pm$ 0.8 & \underline{63.6 $\pm$ 0.7} & 59.7 $\pm$ 1.0 & 53.4 $\pm$ 2.4 & 74.5 $\pm$ 2.5 & 74.6 $\pm$ 1.5 & 70.8 $\pm$ 3.3 \\
   \cdashline{2-10}[1pt/1pt]
          & MoE & 66.9 $\pm$ 2.0 & 74.5 $\pm$ 0.4 & 62.9 $\pm$ 0.7 & 61.7 $\pm$ 1.3 & 60.3 $\pm$ 4.4 & 73.0 $\pm$ 1.6 & 73.5 $\pm$ 1.0 & 71.1 $\pm$ 3.1 \\
          & E-Ensemble & 67.3 $\pm$ 1.6 & 74.5 $\pm$ 0.5 & 62.4 $\pm$ 0.6 & 59.6 $\pm$ 0.8 & \underline{60.9 $\pm$ 3.2} & 73.6 $\pm$ 6.1 & 75.0 $\pm$ 1.1 & 70.1 $\pm$ 2.5 \\
          & GraphDIVE & 61.3 $\pm$ 2.3 & \underline{74.6 $\pm$ 0.5} & 62.3 $\pm$ 0.6 & 57.1 $\pm$ 2.8 & 57.1 $\pm$ 7.6 & 68.2 $\pm$ 3.9 & 65.2 $\pm$ 2.8 & 65.9 $\pm$ 5.2 \\
          & TopExpert & 67.6 $\pm$ 2.0 & 74.3 $\pm$ 0.5 & 62.6 $\pm$ 0.7 & \underline{62.6 $\pm$ 0.9} & 58.7 $\pm$ 2.7 & \underline{76.0 $\pm$ 1.5} & \underline{75.5 $\pm$ 1.0} & \underline{74.0 $\pm$ 3.1} \\
    \cdashline{2-10}[1pt/1pt]
          & \textbf{ASE-Mol} & \textbf{83.6 $\pm$ 1.5} & \textbf{77.0 $\pm$ 0.6} & \textbf{71.7 $\pm$ 0.5} & \textbf{63.3 $\pm$ 0.9} & \textbf{87.9 $\pm$ 3.3} & \textbf{80.4 $\pm$ 2.0} & \textbf{84.3 $\pm$ 1.6} & \textbf{94.8 $\pm$ 2.7} \\
    \bottomrule
    \end{tabular}%
    }
    \label{tab: result}
\end{table*}

\subsubsection{Implementation Details}

Since all eight datasets are related to classification tasks, we use the Receiver Operating Characteristic Area Under the Curve (ROC-AUC) to evaluate the performance of all methods. We performed ten runs with different random seeds and reported the mean and standard deviations. Specifically, our experiments were conducted on an Ubuntu Server equipped with an Intel(R) Core(TM) i7-8700K CPU and two NVIDIA GeForce GTX 1080 Ti GPUs (each with 11GB of memory). Our model was developed and tested in Python 3.7.1, using PyTorch 1.13.1 and PyTorch Geometric 2.3.1.

Following previous works~\cite{kim2023learning}, we set the number of model layers to 5 and the embedding dimension $d$ to 300. The models were trained for a maximum of 100 epochs in the process of substructure recognition optimization and a maximum of 200 epochs in the process of MoE module optimization. Other hyperparameters are obtained by grid search, the search range is as follows: the batch size is \{128, 256, 512\}, the learning rate is \{0.0005, 0.001, 0.005\}, the weight decay is \{0, 1e-5, 1e-4\}, the number of experts is \{1, 3, 5, 7, 9\} for each task, the loss balance parameters $\alpha$ and $\beta$ are \{0.01, 0.1, 1, 5\}, and the parameter $\psi$ that controls the number of substructures is \{0.1, 0.2, 0.3\}.

\begin{table}[!t]
    \caption{Ablation experiment results on the SIDER and HIV dataset.}
    \centering
    \renewcommand{\arraystretch}{1.05}
    \resizebox{0.45\textwidth}{!}{
    \begin{tabular}{rll}
    \toprule
    \multicolumn{1}{c}{\textbf{Methods}} & \multicolumn{1}{c}{\textbf{SIDER}} & \multicolumn{1}{c}{\textbf{HIV}} \\
    \midrule
    \multicolumn{1}{c}{(+GCN)} & \textbf{64.0 $\pm$ 1.2} & \textbf{85.5 $\pm$ 1.1} \\
    w/o N & 60.2 $\pm$ 1.8 \textcolor{black}{ (-3.8)} & 79.6 $\pm$ 3.0 \textcolor{black}{ (-5.9)} \\
    w/o P & 60.1 $\pm$ 1.6 \textcolor{black}{ (-3.9)} & 78.7 $\pm$ 3.6 \textcolor{black}{ (-6.8)} \\
    w/o SLO & 55.3 $\pm$ 1.4 \textcolor{black}{ (-8.7)} & 68.2 $\pm$ 2.7 \textcolor{black}{ (-17.3)} \\
    \midrule
    \multicolumn{1}{c}{(+GIN)} & \textbf{66.5 $\pm$ 1.1} & \textbf{81.9 $\pm$ 1.7} \\
    w/o N & 63.6 $\pm$ 2.8 \textcolor{black}{ (-2.9)} & 78.5 $\pm$ 4.3 \textcolor{black}{ (-3.4)} \\
    w/o P & 60.8 $\pm$ 2.3 \textcolor{black}{ (-5.7)} & 75.8 $\pm$ 2.3 \textcolor{black}{ (-6.1)} \\
    w/o SLO & 55.8 $\pm$ 2.6 \textcolor{black}{ (-10.7)} & 68.9 $\pm$ 3.4 \textcolor{black}{ (-13.0)} \\
    \midrule
    \multicolumn{1}{c}{(+GAT)} & \textbf{63.6 $\pm$ 1.1} & \textbf{86.3 $\pm$ 2.3} \\
    w/o N & 60.6 $\pm$ 2.3 \textcolor{black}{ (-3.0)} & 77.7 $\pm$ 5.1 \textcolor{black}{ (-8.6)} \\
    w/o P & 61.3 $\pm$ 1.3 \textcolor{black}{ (-2.3)} & 80.8 $\pm$ 1.1 \textcolor{black}{ (-5.5)} \\
    w/o SLO & 57.7 $\pm$ 2.4 \textcolor{black}{ (-5.9)} & 71.6 $\pm$ 2.3 \textcolor{black}{ (-14.7)} \\
    \midrule
    \multicolumn{1}{c}{(+GraphSAGE)} & \textbf{63.3 $\pm$ 0.9} & \textbf{84.3 $\pm$ 1.6} \\
    w/o N & 59.5 $\pm$ 1.8 \textcolor{black}{ (-3.8)} & 77.6 $\pm$ 3.6 \textcolor{black}{ (-6.7)} \\
    w/o P & 59.0 $\pm$ 1.2 \textcolor{black}{ (-4.3)} & 77.1 $\pm$ 3.4 \textcolor{black}{ (-7.2)} \\
    w/o SLO & 56.6 $\pm$ 2.1 \textcolor{black}{ (-6.7)} & 69.6 $\pm$ 2.4 \textcolor{black}{ (-14.7)} \\
    \bottomrule
    \label{tab: ablation}
    \end{tabular}
    }
\end{table}

\subsection{Main Results}

To evaluate the effectiveness of ASE-Mol, we conducted comprehensive experiments on eight benchmark datasets, as detailed in~\cref{tab: result}. ASE-Mol consistently outperformed existing methods across all GNN backbones and datasets, with particularly notable improvements on ClinTox and BACE. For instance, on the BACE dataset with a GCN encoder, ASE-Mol achieved a ROC-AUC of 96.3\%, marking an absolute improvement of 26.0\% over the best baseline. Similarly, on the ToxCast dataset, where other MoE-based methods often showed negligible or even adverse effects, ASE-Mol maintained consistently high ROC-AUC values. These improvements can be attributed to ASE-Mol’s ability to effectively distinguish and adapt to different substructures, amplifying the influence of positive motifs while mitigating the impact of negative ones. Moreover, the standard deviation of results across all datasets was comparable to or slightly lower than those of existing methods, demonstrating the robustness of ASE-Mol. Overall, these results highlight that ASE-Mol reinforces its effectiveness in molecular property prediction through interpretable substructure analysis and MoE-based adaptation.

\subsection{Ablation Study}

To investigate the contributions of each component in ASE-Mol, we conducted ablation studies by comparing the full model with three variants: w/o P (ASE-Mol without the ``positive motif''), w/o N (without the ``negative motif''), and w/o SLO (without substructure learning optimization). The results on the SIDER and HIV datasets are illustrated in~\cref{tab: ablation}. We observed that ASE-Mol consistently achieved the best performance when all three components were included. Specifically, for the SIDER dataset, removing the ``positive motif'' (w/o P) led to significant reductions in ROC-AUC scores across all GNN backbones. A similar trend was observed when the ``negative motif'' was excluded (w/o N), highlighting the critical roles of both motifs in capturing relevant molecular information. Moreover, the removal of substructure learning optimization (w/o SLO) also resulted in a notable performance drop, suggesting that this optimization is essential for enhancing the model's ability to learn discriminative substructures. On the HIV dataset, the results further reinforced the importance of these components, with the full ASE-Mol model outperforming all ablation variants across all GNN backbones. 

\begin{figure}[!t]
    \centering
    \includegraphics[width=\linewidth]{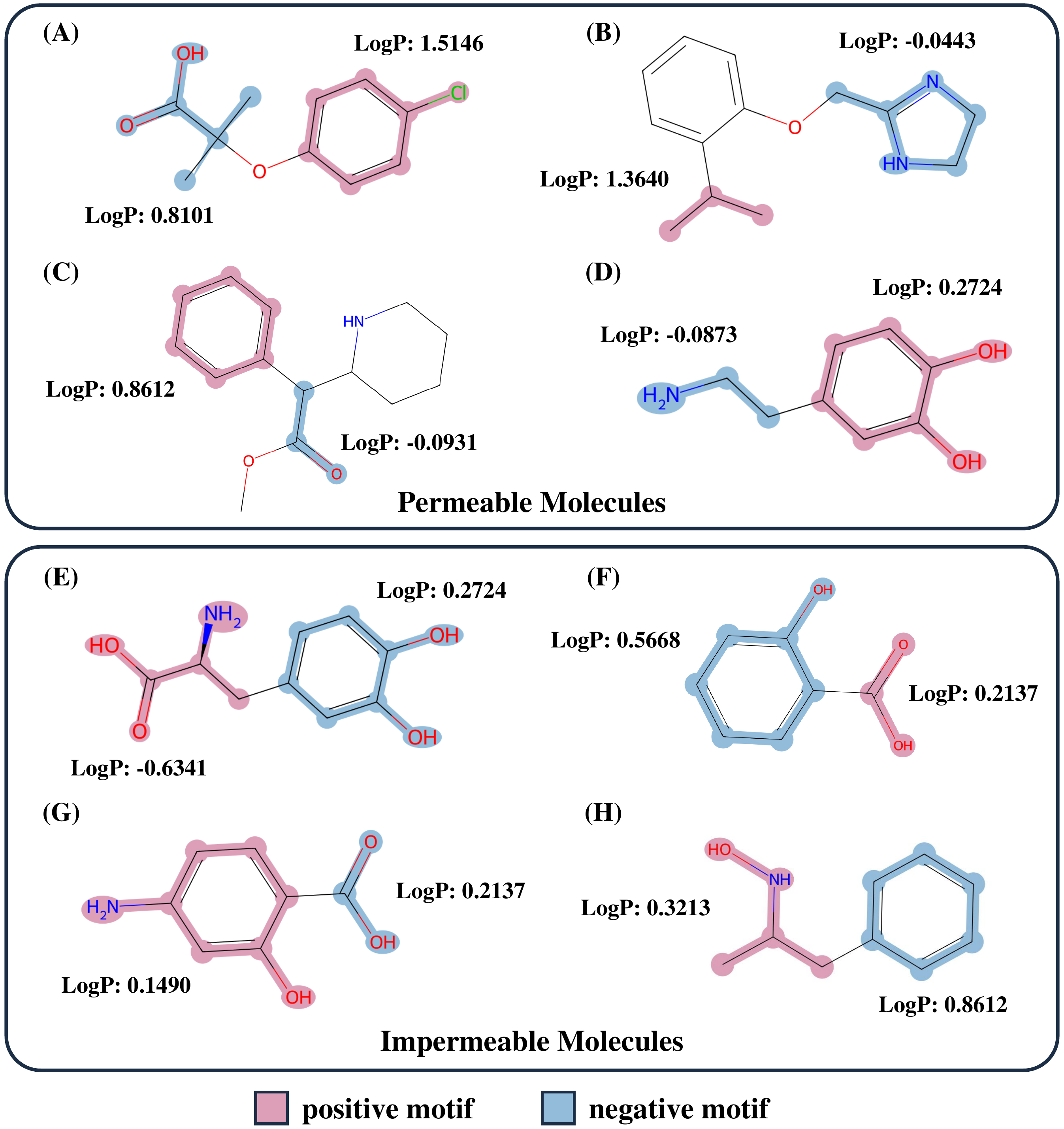}
    \caption{The substructure attribution visualization on the BBBP dataset.}
    \label{fig: visual}
\end{figure}

\subsection{Substructures Attribution Analysis}

\begin{figure*}[!t]
    \centering
    \includegraphics[width=\linewidth]{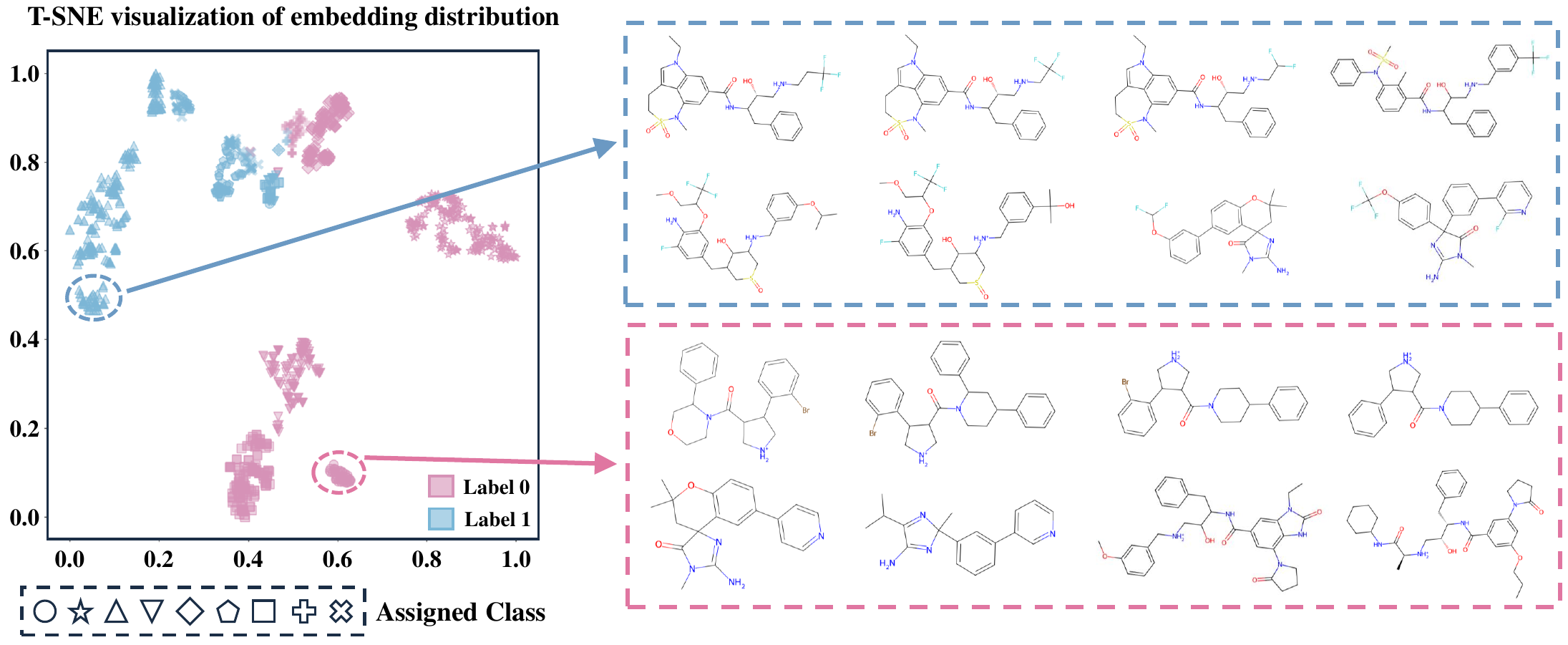}
    \caption{T-SNE visualization with the router assignment on the BACE dataset.}
    \label{fig: assignment}
\end{figure*}

To further explore the relationship between molecular substructures and their properties, we conducted attribution visualization on the BBBP dataset to identify the most influential positive and negative motifs contributing to molecular permeability across the blood-brain barrier (BBB). Molecular permeability is primarily governed by polarity and hydrophobicity, with hydrophobic molecules (low polarity and high LogP values) being more likely to cross the BBB, while hydrophilic molecules (high polarity and low LogP values) exhibit limited permeability~\cite{li2023fg}. As shown in~\cref{fig: visual}, we visualized the attribution weights of molecular substructures, where red indicates positive motifs that support task prediction and blue represents negative motifs that hinder task prediction. That is, in permeable molecules, the positive motifs denote substructures that favor permeability, while in impermeable molecules, the positive motifs denote substructures that hinder permeability. For instance, in the permeable molecule illustrated in~\cref{fig: visual} (C), the phenyl group is identified as the primary contributor to BBB permeability, while the carbonyl with carbon negatively impacts permeability. To further validate these findings, we quantified the LogP values of these substructures using the RDKit~\footnote{https://www.rdkit.org/} open-source toolkit. The former exhibits a LogP value of 0.8612, whereas the latter has a LogP value of -0.0931, which aligns well with the predictions of ASE-Mol. Similarly, ASE-Mol effectively identifies hydrophilic and hydrophobic regions in impermeable molecules. For example, in~\cref{fig: visual} (F), the carboxyl group, highlighted in red, has a LogP value of 0.2137, indicating its hydrophilic nature and its potential role in restricting BBB permeability. By identifying significant motifs in molecules and calculating their LogP values, it can be demonstrated that ASE-Mol can enhance interpretability in addition to improving the performance of molecular property prediction tasks.

\begin{figure*}[ht]
    \centering
    \includegraphics[width=\linewidth]{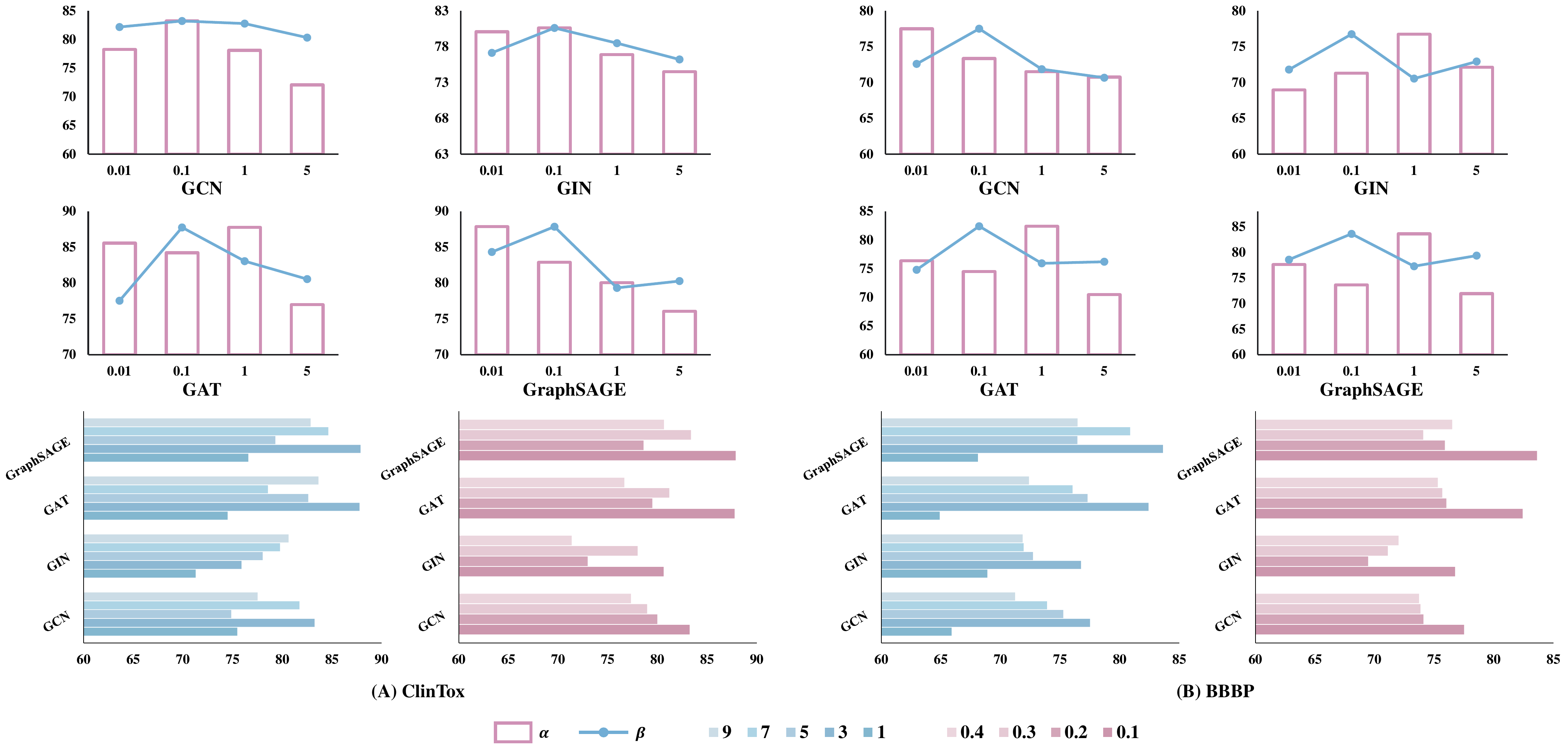}
    \caption{The hyper-parameter sensitivity analysis on the ClinTox and BBBP dataset.}
    \label{fig: parameter}
\end{figure*}

\begin{table*}[!t]
    \centering
    \caption{Detailed time \& memory of ASE-Mol and other baselines on three datasets.}
    \renewcommand{\arraystretch}{1.1}
    \resizebox{0.94\textwidth}{!}{
    \begin{tabular}{cccccccc}
        \toprule
        \multirow{2}{*}{\textbf{Encoder}} & \multirow{2}{*}{\textbf{Method}} & \multicolumn{2}{c}{\textbf{BACE}} & \multicolumn{2}{c}{\textbf{SIDER}} & \multicolumn{2}{c}{\textbf{HIV}} \\
        \cmidrule(r){3-4} \cmidrule(r){5-6} \cmidrule(r){7-8}
        & & Time (ms) & Memory (MB) & Time (ms) & Memory (MB) & Time (ms) & Memory (MB) \\
        \midrule
        \multirow{2}{*}{\textbf{GCN}} & \textbf{ASE-Mol} & 1.43 & 834 & 1.15 & 952 & 7.77 & 2430 \\
                                      & \textbf{TopExpert} & 1.17 & 922 & 1.10 & 1088 & 6.04 & 1944 \\
        \midrule
        \multirow{2}{*}{\textbf{GIN}} & \textbf{ASE-Mol} & 1.31 & 946 & 1.20 & 1040 & 8.82 & 2068 \\
                                      & \textbf{TopExpert} & 1.33 & 1120 & 1.04 & 1244 & 7.31 & 2270 \\
        \midrule
        \multirow{2}{*}{\textbf{GAT}} & \textbf{ASE-Mol} & 1.10 & 1998 & 1.46 & 2868 & 14.73 & 5210 \\
                                      & \textbf{TopExpert} & 1.70 & 2328 & 1.33 & 3114 & 15.50 & 10808 \\
        \midrule
        \multirow{2}{*}{\textbf{GraphSAGE}} & \textbf{ASE-Mol} & 1.40 & 834 & 1.13 & 976 & 8.00 & 2024 \\
                                            & \textbf{TopExpert} & 1.25 & 972 & 1.11 & 1052 & 6.12 & 1956 \\
        \bottomrule
    \label{tab: Time_Complexity}
    \end{tabular}
    }
\end{table*}

\subsection{Router Assignment Visualization}


To further validate the capability of ASE-Mol, we conducted a qualitative analysis using router assignment visualization. Specifically, we applied T-SNE~\cite{van2008visualizing} to reduce the dimensionality of molecular embeddings for visualization. The visualization result for the BACE dataset is presented in~\cref{fig: assignment}. In this visualization, the assigned classes are formed by the pairwise combinations of two types of three experts, representing positive and negative motif experts, respectively. The result shows that molecular embeddings cluster distinctly according to the dataset labels, demonstrating the effectiveness of ASE-Mol in classification tasks. The clear separation of the routing-assigned clusters suggests that the embeddings generated by our method successfully capture the essential features needed to distinguish molecular properties. Furthermore, to assess the interpretability of the model, we visualized the specified molecules routed to the same cluster. By examining the molecules assigned to the same class (e.g., circled in blue and red), we observed notable similarities in their BRICS fragments. This alignment between molecular embeddings and functional group patterns indicates that the model not only achieves accurate classification but also effectively captures domain-relevant chemical characteristics, reinforcing its interpretability.

\subsection{Hyper-parameter sensitivity}

To investigate the sensitivity of ASE-Mol to hyper-parameters, we analyzed the impact of four key factors: the number of experts, the parameter $\psi$ controlling the number of substructures, and the loss balance parameters $\alpha$ and $\beta$. \cref{fig: parameter} illustrates the performance across different settings of these parameters. First, for the loss balance parameters $\alpha$ and $\beta$, a proper combination leads to better performance across various GNN backbones, while an inappropriate ratio between $\alpha$ and $\beta$ can cause a decline in performance, indicating the importance of balancing different loss terms. Regarding the number of experts, the optimal value varied depending on the dataset and GNN backbone used. This suggests that the number of experts should be tuned using a validation set for each scenario. Moreover, the parameter $\psi$ has a significant impact on performance. On the ClinTox and BBBP datasets, excessively large $\psi$ values introduce redundant or noisy substructures that reduce the effectiveness of the model. In conclusion, careful tuning of these hyperparameters is crucial for maximizing the performance of ASE-Mol.

\subsection{Time Complexity}

To conduct a comprehensive performance evaluation, we compared the time complexity of ASE-Mol with TopExpert across multiple datasets and GNN backbones. Specifically, we selected three datasets (BACE, SIDER, and HIV) with different scales and recorded the per-epoch training time and GPU memory usage. The experimental results are presented in~\cref{tab: Time_Complexity}. To ensure fair comparisons, all experiments were conducted using the same batch size. The results show that ASE-Mol achieves similar running time and GPU memory usage to the TopExpert for most datasets and backbones. Notably, ASE-Mol demonstrates significantly lower memory usage on GAT, highlighting its efficiency. Moreover, it achieves substantial performance improvements across all GNN backbones, validating its effectiveness and practicality.

\section{Conclusion}\label{sec:conclusion}

In this work, we introduce ASE-Mol, a novel framework that leverages a Mixture-of-Experts architecture to enhance molecular property prediction. By combining BRICS decomposition with significant substructure awareness, ASE-Mol effectively identifies and distinguishes between positive and negative motifs, dynamically routing molecular representations to specialized experts. The framework demonstrates significant performance improvements across eight benchmark datasets, consistently outperforming state-of-the-art methods. Furthermore, interpretability analysis reveals that the framework can accurately pinpoint key substructures influencing molecular properties, providing valuable insights for molecular property prediction tasks. However, as ASE-Mol utilizes positive and negative motifs, it currently focuses on classification tasks. This limitation restricts its applicability to regression tasks, where the target values are continuous. Future work will aim to extend the framework’s capabilities to regression scenarios, thereby broadening its potential applications in molecular property prediction.




\bibliographystyle{elsarticle-num-names} 
\bibliography{elsarticle-template-num-names}

\begin{thebibliography}{40}
\expandafter\ifx\csname natexlab\endcsname\relax\def\natexlab#1{#1}\fi
\providecommand{\url}[1]{\texttt{#1}}
\providecommand{\href}[2]{#2}
\providecommand{\path}[1]{#1}
\providecommand{\DOIprefix}{doi:}
\providecommand{\ArXivprefix}{arXiv:}
\providecommand{\URLprefix}{URL: }
\providecommand{\Pubmedprefix}{pmid:}
\providecommand{\doi}[1]{\href{http://dx.doi.org/#1}{\path{#1}}}
\providecommand{\Pubmed}[1]{\href{pmid:#1}{\path{#1}}}
\providecommand{\bibinfo}[2]{#2}
\ifx\xfnm\relax \def\xfnm[#1]{\unskip,\space#1}\fi
\bibitem[{Walters and Barzilay(2020)}]{walters2020applications}
\bibinfo{author}{W.~P. Walters}, \bibinfo{author}{R.~Barzilay},
\newblock \bibinfo{title}{Applications of deep learning in molecule generation and molecular property prediction},
\newblock \bibinfo{journal}{Accounts of chemical research} \bibinfo{volume}{54} (\bibinfo{year}{2020}) \bibinfo{pages}{263--270}.
\bibitem[{Wang et~al.(2024{\natexlab{a}})Wang, Jiang, Wang, and Xuan}]{wang2024multi}
\bibinfo{author}{Z.~Wang}, \bibinfo{author}{T.~Jiang}, \bibinfo{author}{J.~Wang}, \bibinfo{author}{Q.~Xuan}, \bibinfo{title}{Multi-modal representation learning for molecular property prediction: Sequence, graph, geometry}, \bibinfo{year}{2024}{\natexlab{a}}. \URLprefix \url{https://arxiv.org/abs/2401.03369}. \href{http://arxiv.org/abs/2401.03369}{{\tt arXiv:2401.03369}}.
\bibitem[{Wang et~al.(2024{\natexlab{b}})Wang, Shao, Wang, Yu, Xuan, and Yang}]{wang2024sub}
\bibinfo{author}{J.~Wang}, \bibinfo{author}{J.~Shao}, \bibinfo{author}{Z.~Wang}, \bibinfo{author}{S.~Yu}, \bibinfo{author}{Q.~Xuan}, \bibinfo{author}{X.~Yang}, \bibinfo{title}{Subgraph networks based contrastive learning}, \bibinfo{year}{2024}{\natexlab{b}}. \URLprefix \url{https://arxiv.org/abs/2306.03506}. \href{http://arxiv.org/abs/2306.03506}{{\tt arXiv:2306.03506}}.
\bibitem[{Jiang et~al.(2024)Jiang, Wang, Yu, Wang, Yu, Bao, Wei, and Xuan}]{jiang2024mix}
\bibinfo{author}{T.~Jiang}, \bibinfo{author}{Z.~Wang}, \bibinfo{author}{W.~Yu}, \bibinfo{author}{J.~Wang}, \bibinfo{author}{S.~Yu}, \bibinfo{author}{X.~Bao}, \bibinfo{author}{B.~Wei}, \bibinfo{author}{Q.~Xuan},
\newblock \bibinfo{title}{Mix-key: graph mixup with key structures for molecular property prediction},
\newblock \bibinfo{journal}{Briefings in Bioinformatics} \bibinfo{volume}{25} (\bibinfo{year}{2024}) \bibinfo{pages}{bbae165}.
\bibitem[{Wang et~al.(2024)Wang, Jiang, Lu, Bao, Yu, Wei, and Xuan}]{wang2024know}
\bibinfo{author}{Z.~Wang}, \bibinfo{author}{T.~Jiang}, \bibinfo{author}{Y.~Lu}, \bibinfo{author}{X.~Bao}, \bibinfo{author}{S.~Yu}, \bibinfo{author}{B.~Wei}, \bibinfo{author}{Q.~Xuan}, \bibinfo{title}{Knowledge-enhanced relation graph and task sampling for few-shot molecular property prediction}, \bibinfo{year}{2024}. \URLprefix \url{https://arxiv.org/abs/2405.15544}. \href{http://arxiv.org/abs/2405.15544}{{\tt arXiv:2405.15544}}.
\bibitem[{Wang et~al.(2023)Wang, Wang, Shan, Yu, Xu, Xuan, and Chen}]{wang2023null}
\bibinfo{author}{Z.~Wang}, \bibinfo{author}{J.~Wang}, \bibinfo{author}{Y.~Shan}, \bibinfo{author}{S.~Yu}, \bibinfo{author}{X.~Xu}, \bibinfo{author}{Q.~Xuan}, \bibinfo{author}{G.~Chen},
\newblock \bibinfo{title}{Null model-based data augmentation for graph classification},
\newblock \bibinfo{journal}{IEEE Transactions on Network Science and Engineering}  (\bibinfo{year}{2023}).
\bibitem[{Yuksel et~al.(2012)Yuksel, Wilson, and Gader}]{6215056}
\bibinfo{author}{S.~E. Yuksel}, \bibinfo{author}{J.~N. Wilson}, \bibinfo{author}{P.~D. Gader},
\newblock \bibinfo{title}{Twenty years of mixture of experts},
\newblock \bibinfo{journal}{IEEE Transactions on Neural Networks and Learning Systems} \bibinfo{volume}{23} (\bibinfo{year}{2012}) \bibinfo{pages}{1177--1193}. \DOIprefix\doi{10.1109/TNNLS.2012.2200299}.
\bibitem[{Hu et~al.(2022)Hu, Wang, Liu, Wu, Wang, and Tan}]{ijcai2022p289}
\bibinfo{author}{F.~Hu}, \bibinfo{author}{L.~Wang}, \bibinfo{author}{Q.~Liu}, \bibinfo{author}{S.~Wu}, \bibinfo{author}{L.~Wang}, \bibinfo{author}{T.~Tan},
\newblock \bibinfo{title}{Graphdive: Graph classification by mixture of diverse experts},
\newblock in: \bibinfo{editor}{L.~D. Raedt} (Ed.), \bibinfo{booktitle}{Proceedings of the Thirty-First International Joint Conference on Artificial Intelligence, {IJCAI-22}}, \bibinfo{publisher}{International Joint Conferences on Artificial Intelligence Organization}, \bibinfo{year}{2022}, pp. \bibinfo{pages}{2080--2086}. \URLprefix \url{https://doi.org/10.24963/ijcai.2022/289}. \DOIprefix\doi{10.24963/ijcai.2022/289}, \bibinfo{note}{main Track}.
\bibitem[{Wang et~al.(2024)Wang, Jiang, You, Han, Liu, Srinivasa, Kompella, Wang et~al.}]{wang2024graph}
\bibinfo{author}{H.~Wang}, \bibinfo{author}{Z.~Jiang}, \bibinfo{author}{Y.~You}, \bibinfo{author}{Y.~Han}, \bibinfo{author}{G.~Liu}, \bibinfo{author}{J.~Srinivasa}, \bibinfo{author}{R.~Kompella}, \bibinfo{author}{Z.~Wang}, et~al.,
\newblock \bibinfo{title}{Graph mixture of experts: Learning on large-scale graphs with explicit diversity modeling},
\newblock \bibinfo{journal}{Advances in Neural Information Processing Systems} \bibinfo{volume}{36} (\bibinfo{year}{2024}).
\bibitem[{Kim et~al.(2023)Kim, Lee, Kang, Lee, and Yu}]{kim2023learning}
\bibinfo{author}{S.~Kim}, \bibinfo{author}{D.~Lee}, \bibinfo{author}{S.~Kang}, \bibinfo{author}{S.~Lee}, \bibinfo{author}{H.~Yu},
\newblock \bibinfo{title}{Learning topology-specific experts for molecular property prediction},
\newblock in: \bibinfo{booktitle}{Proceedings of the AAAI Conference on Artificial Intelligence}, volume~\bibinfo{volume}{37}, \bibinfo{year}{2023}, pp. \bibinfo{pages}{8291--8299}.
\bibitem[{Degen et~al.(2008)Degen, Wegscheid-Gerlach, Zaliani, and Rarey}]{degen2008art}
\bibinfo{author}{J.~Degen}, \bibinfo{author}{C.~Wegscheid-Gerlach}, \bibinfo{author}{A.~Zaliani}, \bibinfo{author}{M.~Rarey},
\newblock \bibinfo{title}{On the art of compiling and using'drug-like'chemical fragment spaces},
\newblock \bibinfo{journal}{ChemMedChem} \bibinfo{volume}{3} (\bibinfo{year}{2008}) \bibinfo{pages}{1503}.
\bibitem[{Li et~al.(2022)Li, Jiang, Wang, and Zhang}]{li2022deep}
\bibinfo{author}{Z.~Li}, \bibinfo{author}{M.~Jiang}, \bibinfo{author}{S.~Wang}, \bibinfo{author}{S.~Zhang},
\newblock \bibinfo{title}{Deep learning methods for molecular representation and property prediction},
\newblock \bibinfo{journal}{Drug Discovery Today} \bibinfo{volume}{27} (\bibinfo{year}{2022}) \bibinfo{pages}{103373}.
\bibitem[{Xia et~al.(2023)Xia, Zhang, Zhu, Liu, Gao, Hu, Tan, Zheng, Li, and Li}]{xia2023understanding}
\bibinfo{author}{J.~Xia}, \bibinfo{author}{L.~Zhang}, \bibinfo{author}{X.~Zhu}, \bibinfo{author}{Y.~Liu}, \bibinfo{author}{Z.~Gao}, \bibinfo{author}{B.~Hu}, \bibinfo{author}{C.~Tan}, \bibinfo{author}{J.~Zheng}, \bibinfo{author}{S.~Li}, \bibinfo{author}{S.~Z. Li},
\newblock \bibinfo{title}{Understanding the limitations of deep models for molecular property prediction: Insights and solutions},
\newblock \bibinfo{journal}{Advances in Neural Information Processing Systems} \bibinfo{volume}{36} (\bibinfo{year}{2023}) \bibinfo{pages}{64774--64792}.
\bibitem[{Xuan et~al.(2019)Xuan, Wang, Zhao, Yuan, Fu, Ruan, and Chen}]{xuan2019subgraph}
\bibinfo{author}{Q.~Xuan}, \bibinfo{author}{J.~Wang}, \bibinfo{author}{M.~Zhao}, \bibinfo{author}{J.~Yuan}, \bibinfo{author}{C.~Fu}, \bibinfo{author}{Z.~Ruan}, \bibinfo{author}{G.~Chen},
\newblock \bibinfo{title}{Subgraph networks with application to structural feature space expansion},
\newblock \bibinfo{journal}{IEEE Transactions on Knowledge and Data Engineering} \bibinfo{volume}{33} (\bibinfo{year}{2019}) \bibinfo{pages}{2776--2789}.
\bibitem[{Zhou et~al.(2020)Zhou, Shen, Yu, Chen, and Xuan}]{zhou2020m}
\bibinfo{author}{J.~Zhou}, \bibinfo{author}{J.~Shen}, \bibinfo{author}{S.~Yu}, \bibinfo{author}{G.~Chen}, \bibinfo{author}{Q.~Xuan},
\newblock \bibinfo{title}{M-evolve: structural-mapping-based data augmentation for graph classification},
\newblock \bibinfo{journal}{IEEE Transactions on Network Science and Engineering} \bibinfo{volume}{8} (\bibinfo{year}{2020}) \bibinfo{pages}{190--200}.
\bibitem[{Yang et~al.(2024)Yang, Zhang, Wu, Yang, Sheng, Xue, Zhou, Aggarwal, Peng, Hu et~al.}]{yang2024state}
\bibinfo{author}{Z.~Yang}, \bibinfo{author}{G.~Zhang}, \bibinfo{author}{J.~Wu}, \bibinfo{author}{J.~Yang}, \bibinfo{author}{Q.~Z. Sheng}, \bibinfo{author}{S.~Xue}, \bibinfo{author}{C.~Zhou}, \bibinfo{author}{C.~Aggarwal}, \bibinfo{author}{H.~Peng}, \bibinfo{author}{W.~Hu}, et~al.,
\newblock \bibinfo{title}{State of the art and potentialities of graph-level learning},
\newblock \bibinfo{journal}{ACM Computing Surveys} \bibinfo{volume}{57} (\bibinfo{year}{2024}) \bibinfo{pages}{1--40}.
\bibitem[{Zhang et~al.(2021)Zhang, Liu, Wang, Lu, and Lee}]{zhang2021motif}
\bibinfo{author}{Z.~Zhang}, \bibinfo{author}{Q.~Liu}, \bibinfo{author}{H.~Wang}, \bibinfo{author}{C.~Lu}, \bibinfo{author}{C.-K. Lee},
\newblock \bibinfo{title}{Motif-based graph self-supervised learning for molecular property prediction},
\newblock \bibinfo{journal}{Advances in Neural Information Processing Systems} \bibinfo{volume}{34} (\bibinfo{year}{2021}) \bibinfo{pages}{15870--15882}.
\bibitem[{Xie et~al.(2023)Xie, Zhang, Guan, and Zhou}]{xie2023self}
\bibinfo{author}{A.~Xie}, \bibinfo{author}{Z.~Zhang}, \bibinfo{author}{J.~Guan}, \bibinfo{author}{S.~Zhou},
\newblock \bibinfo{title}{Self-supervised learning with chemistry-aware fragmentation for effective molecular property prediction},
\newblock \bibinfo{journal}{Briefings in Bioinformatics} \bibinfo{volume}{24} (\bibinfo{year}{2023}) \bibinfo{pages}{bbad296}.
\bibitem[{Liu et~al.(2024)Liu, Shi, Zhang, Zhang, Kawaguchi, Wang, and Chua}]{liu2024rethinking}
\bibinfo{author}{Z.~Liu}, \bibinfo{author}{Y.~Shi}, \bibinfo{author}{A.~Zhang}, \bibinfo{author}{E.~Zhang}, \bibinfo{author}{K.~Kawaguchi}, \bibinfo{author}{X.~Wang}, \bibinfo{author}{T.-S. Chua},
\newblock \bibinfo{title}{Rethinking tokenizer and decoder in masked graph modeling for molecules},
\newblock \bibinfo{journal}{Advances in Neural Information Processing Systems} \bibinfo{volume}{36} (\bibinfo{year}{2024}).
\bibitem[{Shazeer et~al.(2017)Shazeer, Mirhoseini, Maziarz, Davis, Le, Hinton, and Dean}]{shazeer2017}
\bibinfo{author}{N.~Shazeer}, \bibinfo{author}{A.~Mirhoseini}, \bibinfo{author}{K.~Maziarz}, \bibinfo{author}{A.~Davis}, \bibinfo{author}{Q.~Le}, \bibinfo{author}{G.~Hinton}, \bibinfo{author}{J.~Dean},
\newblock \bibinfo{title}{Outrageously large neural networks: The sparsely-gated mixture-of-experts layer},
\newblock in: \bibinfo{booktitle}{International Conference on Learning Representations}, \bibinfo{year}{2017}. \URLprefix \url{https://openreview.net/forum?id=B1ckMDqlg}.
\bibitem[{Jacobs et~al.(1991)Jacobs, Jordan, Nowlan, and Hinton}]{jacobs1991adaptive}
\bibinfo{author}{R.~A. Jacobs}, \bibinfo{author}{M.~I. Jordan}, \bibinfo{author}{S.~J. Nowlan}, \bibinfo{author}{G.~E. Hinton},
\newblock \bibinfo{title}{Adaptive mixtures of local experts},
\newblock \bibinfo{journal}{Neural computation} \bibinfo{volume}{3} (\bibinfo{year}{1991}) \bibinfo{pages}{79--87}.
\bibitem[{Ma et~al.(2018)Ma, Zhao, Yi, Chen, Hong, and Chi}]{ma2018modeling}
\bibinfo{author}{J.~Ma}, \bibinfo{author}{Z.~Zhao}, \bibinfo{author}{X.~Yi}, \bibinfo{author}{J.~Chen}, \bibinfo{author}{L.~Hong}, \bibinfo{author}{E.~H. Chi},
\newblock \bibinfo{title}{Modeling task relationships in multi-task learning with multi-gate mixture-of-experts},
\newblock in: \bibinfo{booktitle}{Proceedings of the 24th ACM SIGKDD international conference on knowledge discovery \& data mining}, \bibinfo{year}{2018}, pp. \bibinfo{pages}{1930--1939}.
\bibitem[{Zhang et~al.(2024)Zhang, Sun, Yue, Jiang, Wang, Chen, and Pan}]{zhang2024graph}
\bibinfo{author}{G.~Zhang}, \bibinfo{author}{X.~Sun}, \bibinfo{author}{Y.~Yue}, \bibinfo{author}{C.~Jiang}, \bibinfo{author}{K.~Wang}, \bibinfo{author}{T.~Chen}, \bibinfo{author}{S.~Pan},
\newblock \bibinfo{title}{Graph sparsification via mixture of graphs},
\newblock \bibinfo{journal}{arXiv preprint arXiv:2405.14260}  (\bibinfo{year}{2024}).
\bibitem[{Yao et~al.(2024)Yao, Liu, Meng, Zhan, Wu, Pan, and Hu}]{yao2024moe}
\bibinfo{author}{Z.~Yao}, \bibinfo{author}{C.~Liu}, \bibinfo{author}{X.~Meng}, \bibinfo{author}{Y.~Zhan}, \bibinfo{author}{J.~Wu}, \bibinfo{author}{S.~Pan}, \bibinfo{author}{W.~Hu},
\newblock \bibinfo{title}{Da-moe: Addressing depth-sensitivity in graph-level analysis through mixture of experts},
\newblock \bibinfo{journal}{arXiv preprint arXiv:2411.03025}  (\bibinfo{year}{2024}).
\bibitem[{Gaspar and Seddon(2022)}]{gaspar2022glolloc}
\bibinfo{author}{H.~A. Gaspar}, \bibinfo{author}{M.~P. Seddon},
\newblock \bibinfo{title}{Glolloc: Mixture of global and local experts for molecular activity prediction},
\newblock in: \bibinfo{booktitle}{ICLR2022 Machine Learning for Drug Discovery}, \bibinfo{year}{2022}. \URLprefix \url{https://openreview.net/forum?id=Mdj229oYWa3}.
\bibitem[{Wu et~al.(2022)Wu, Wang, Wu, Zhang, Deng, Kang, Cao, Hsieh, and Hou}]{wu2022alipsol}
\bibinfo{author}{J.~Wu}, \bibinfo{author}{J.~Wang}, \bibinfo{author}{Z.~Wu}, \bibinfo{author}{S.~Zhang}, \bibinfo{author}{Y.~Deng}, \bibinfo{author}{Y.~Kang}, \bibinfo{author}{D.~Cao}, \bibinfo{author}{C.-Y. Hsieh}, \bibinfo{author}{T.~Hou},
\newblock \bibinfo{title}{Alipsol: an attention-driven mixture-of-experts model for lipophilicity and solubility prediction},
\newblock \bibinfo{journal}{Journal of Chemical Information and Modeling} \bibinfo{volume}{62} (\bibinfo{year}{2022}) \bibinfo{pages}{5975--5987}.
\bibitem[{Zhang et~al.(2024)Zhang, Qian, Xia, and Yang}]{zhang2024mol}
\bibinfo{author}{X.~Zhang}, \bibinfo{author}{C.~Qian}, \bibinfo{author}{J.~Xia}, \bibinfo{author}{F.~Yang},
\newblock \bibinfo{title}{Mol-moe: Learning drug molecular characterization based on mixture of expert mechanism},
\newblock in: \bibinfo{booktitle}{International Symposium on Bioinformatics Research and Applications}, \bibinfo{organization}{Springer}, \bibinfo{year}{2024}, pp. \bibinfo{pages}{233--244}.
\bibitem[{Wu et~al.(2023)Wu, Wang, Du, Jiang, Kang, Li, Pan, Deng, Cao, Hsieh et~al.}]{wu2023chemistry}
\bibinfo{author}{Z.~Wu}, \bibinfo{author}{J.~Wang}, \bibinfo{author}{H.~Du}, \bibinfo{author}{D.~Jiang}, \bibinfo{author}{Y.~Kang}, \bibinfo{author}{D.~Li}, \bibinfo{author}{P.~Pan}, \bibinfo{author}{Y.~Deng}, \bibinfo{author}{D.~Cao}, \bibinfo{author}{C.-Y. Hsieh}, et~al.,
\newblock \bibinfo{title}{Chemistry-intuitive explanation of graph neural networks for molecular property prediction with substructure masking},
\newblock \bibinfo{journal}{Nature Communications} \bibinfo{volume}{14} (\bibinfo{year}{2023}) \bibinfo{pages}{2585}.
\bibitem[{Schroff et~al.(2015)Schroff, Kalenichenko, and Philbin}]{schroff2015facenet}
\bibinfo{author}{F.~Schroff}, \bibinfo{author}{D.~Kalenichenko}, \bibinfo{author}{J.~Philbin},
\newblock \bibinfo{title}{Facenet: A unified embedding for face recognition and clustering},
\newblock in: \bibinfo{booktitle}{Proceedings of the IEEE conference on computer vision and pattern recognition}, \bibinfo{year}{2015}, pp. \bibinfo{pages}{815--823}.
\bibitem[{Riquelme et~al.(2021)Riquelme, Puigcerver, Mustafa, Neumann, Jenatton, Susano~Pinto, Keysers, and Houlsby}]{riquelme2021scaling}
\bibinfo{author}{C.~Riquelme}, \bibinfo{author}{J.~Puigcerver}, \bibinfo{author}{B.~Mustafa}, \bibinfo{author}{M.~Neumann}, \bibinfo{author}{R.~Jenatton}, \bibinfo{author}{A.~Susano~Pinto}, \bibinfo{author}{D.~Keysers}, \bibinfo{author}{N.~Houlsby},
\newblock \bibinfo{title}{Scaling vision with sparse mixture of experts},
\newblock \bibinfo{journal}{Advances in Neural Information Processing Systems} \bibinfo{volume}{34} (\bibinfo{year}{2021}) \bibinfo{pages}{8583--8595}.
\bibitem[{Jiang et~al.(2024)Jiang, Liu, and Chen}]{jiang2024mope}
\bibinfo{author}{R.~Jiang}, \bibinfo{author}{L.~Liu}, \bibinfo{author}{C.~Chen},
\newblock \bibinfo{title}{Mope: Parameter-efficient and scalable multimodal fusion via mixture of prompt experts},
\newblock \bibinfo{journal}{arXiv preprint arXiv:2403.10568}  (\bibinfo{year}{2024}).
\bibitem[{Wu et~al.(2018)Wu, Ramsundar, Feinberg, Gomes, Geniesse, Pappu, Leswing, and Pande}]{wu2018moleculenet}
\bibinfo{author}{Z.~Wu}, \bibinfo{author}{B.~Ramsundar}, \bibinfo{author}{E.~N. Feinberg}, \bibinfo{author}{J.~Gomes}, \bibinfo{author}{C.~Geniesse}, \bibinfo{author}{A.~S. Pappu}, \bibinfo{author}{K.~Leswing}, \bibinfo{author}{V.~Pande},
\newblock \bibinfo{title}{Moleculenet: a benchmark for molecular machine learning},
\newblock \bibinfo{journal}{Chemical science} \bibinfo{volume}{9} (\bibinfo{year}{2018}) \bibinfo{pages}{513--530}.
\bibitem[{Heid et~al.(2023)Heid, Greenman, Chung, Li, Graff, Vermeire, Wu, Green, and McGill}]{heid2023chemprop}
\bibinfo{author}{E.~Heid}, \bibinfo{author}{K.~P. Greenman}, \bibinfo{author}{Y.~Chung}, \bibinfo{author}{S.-C. Li}, \bibinfo{author}{D.~E. Graff}, \bibinfo{author}{F.~H. Vermeire}, \bibinfo{author}{H.~Wu}, \bibinfo{author}{W.~H. Green}, \bibinfo{author}{C.~J. McGill},
\newblock \bibinfo{title}{Chemprop: a machine learning package for chemical property prediction},
\newblock \bibinfo{journal}{Journal of Chemical Information and Modeling} \bibinfo{volume}{64} (\bibinfo{year}{2023}) \bibinfo{pages}{9--17}.
\bibitem[{Kipf and Welling(2017)}]{kipf2017semisupervised}
\bibinfo{author}{T.~N. Kipf}, \bibinfo{author}{M.~Welling},
\newblock \bibinfo{title}{Semi-supervised classification with graph convolutional networks},
\newblock in: \bibinfo{booktitle}{International Conference on Learning Representations}, \bibinfo{year}{2017}. \URLprefix \url{https://openreview.net/forum?id=SJU4ayYgl}.
\bibitem[{Xu et~al.(2019)Xu, Hu, Leskovec, and Jegelka}]{xu2018how}
\bibinfo{author}{K.~Xu}, \bibinfo{author}{W.~Hu}, \bibinfo{author}{J.~Leskovec}, \bibinfo{author}{S.~Jegelka},
\newblock \bibinfo{title}{How powerful are graph neural networks?},
\newblock in: \bibinfo{booktitle}{International Conference on Learning Representations}, \bibinfo{year}{2019}.
\bibitem[{Veli{\v{c}}kovi{\'c} et~al.(2018)Veli{\v{c}}kovi{\'c}, Cucurull, Casanova, Romero, Li{\`o}, and Bengio}]{velivckovic2018graph}
\bibinfo{author}{P.~Veli{\v{c}}kovi{\'c}}, \bibinfo{author}{G.~Cucurull}, \bibinfo{author}{A.~Casanova}, \bibinfo{author}{A.~Romero}, \bibinfo{author}{P.~Li{\`o}}, \bibinfo{author}{Y.~Bengio},
\newblock \bibinfo{title}{Graph attention networks},
\newblock in: \bibinfo{booktitle}{International Conference on Learning Representations}, \bibinfo{year}{2018}.
\bibitem[{Hamilton et~al.(2017)Hamilton, Ying, and Leskovec}]{hamilton2017inductive}
\bibinfo{author}{W.~Hamilton}, \bibinfo{author}{Z.~Ying}, \bibinfo{author}{J.~Leskovec},
\newblock \bibinfo{title}{Inductive representation learning on large graphs},
\newblock \bibinfo{journal}{Advances in neural information processing systems} \bibinfo{volume}{30} (\bibinfo{year}{2017}).
\bibitem[{Dietterich(2000)}]{dietterich2000ensemble}
\bibinfo{author}{T.~G. Dietterich},
\newblock \bibinfo{title}{Ensemble methods in machine learning},
\newblock in: \bibinfo{booktitle}{International workshop on multiple classifier systems}, \bibinfo{organization}{Springer}, \bibinfo{year}{2000}, pp. \bibinfo{pages}{1--15}.
\bibitem[{Li et~al.(2023)Li, Lin, Chen, and Wang}]{li2023fg}
\bibinfo{author}{B.~Li}, \bibinfo{author}{M.~Lin}, \bibinfo{author}{T.~Chen}, \bibinfo{author}{L.~Wang},
\newblock \bibinfo{title}{Fg-bert: a generalized and self-supervised functional group-based molecular representation learning framework for properties prediction},
\newblock \bibinfo{journal}{Briefings in Bioinformatics} \bibinfo{volume}{24} (\bibinfo{year}{2023}) \bibinfo{pages}{bbad398}.
\bibitem[{Van~der Maaten and Hinton(2008)}]{van2008visualizing}
\bibinfo{author}{L.~Van~der Maaten}, \bibinfo{author}{G.~Hinton},
\newblock \bibinfo{title}{Visualizing data using t-sne.},
\newblock \bibinfo{journal}{Journal of machine learning research} \bibinfo{volume}{9} (\bibinfo{year}{2008}).

\end{thebibliography}






\end{document}